# ClaRet - A CNN Architecture for Optical Coherence Tomography

Adit Magotra*, Aagat Gedam*, Tanush Savadi*, Emily Li*

ARQuest SSERN Intl,

## Abstract

Optical Coherence Tomography is a technique used to scan the Retina of the eye and check for tears. In this paper, we develop a Convolutional Neural Network Architecture for OCT scan classification. The model is trained to detect Retinal tears from an OCT scan and classify the type of tear. We designed a block-based approach to accompany a pre-trained VGG-19 using Transfer Learning by writing customised layers in blocks for better feature extraction. The approach achieved substantially better results than the baseline we initially started out with.

## 1. Introduction

An OCT eye exam is a non-invasive test that provides 3-D cross-sectional images of the retina to enable early detection and treatment of ocular disease that may develop without any noticeable symptoms. Humans lack the ability to go through several OCT scans in a short period of time while today's AI can do it in a matter of seconds. This allows the AI to monitor and assess patient health in the near future in a way that humans cannot. With a fast OCT scanner, AI can process in real-time and make faster decisions that will change the treatment outcomes of humans. The AI can also be used during emergencies, while the human specialist can focus on more important things and form a Plan of Action for the concerned patient. AI's ability to analyze data and make decisions based on patterns in data is the point we have taken and made our model flexible enough to be applied to different fields with just a few simple edits. This is where AI can be used. Its first major advantage is the speed it offers while the second advantage is that it can continue working without a break. This would help in reducing the manual workload on humans and help them focus on more important tasks.

## 2. Problem Statement

The generic approach that involves using pre-trained models along with transfer learning gives out considerable results. But when referenced through the code and layer architecture, there are some flaws and areas of improvement. To increase accuracy and reside the neurons in the layers that are not in use, people use max pooling and Dropout layers respectively. This does pose a satisfactory result. But by doing this[1] we are leaving behind some information in terms of parameters and offset values which are just ignored in the training process. Which could keep us away from the best possible results. Alongside this, the use of dense layers and filters lacks coordination. Referencing a lot of codes and research papers, the average transfer learning approach includes gradually decreasing the dense layers but a set of stationary filters. This does seem to alter the best result

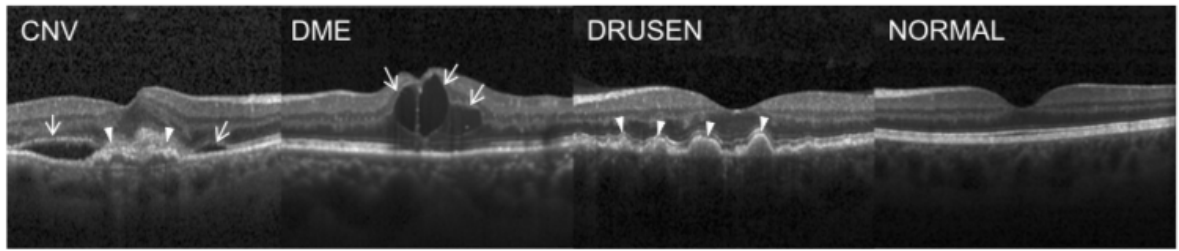

## 2.1 Related works

Eye illness diagnosis and detection are separated into characteristics based on manual segmentation of the vertical cup-to-disc ratio and recognition based on a convolutional neural network (CNN). Full presented a novel integrated optic disc perception network based on global retina pictures and local optic disc regions. developed a new set of characteristics to characterize the morphology of a patient's blood vessels that are independent of visual field or resolution. According to preliminary data, such characteristics can be utilized to evaluate image quality an order of magnitude quicker than previous approaches and methods. presented an artificial eye disease expert system based on segmentation of the optic disc and cup. For the automated diagnosis of eye illness, a CNN-based deep learning architecture was created. proposed a novel normalization approach for extracting four additional features linked to blood vessel illumination reflection, and the features were validated by using the accuracy of a classifier algorithm, i.e. Fisher's linear discriminant, more specifically, the linear discriminant analysis classifier extracted several characteristics from the blood vessel's centre point and used a feature selection technique based on the genetic search to achieve the best feature subset of blood vessel classification.

## 2.2 Previous work:

Deep CNNs have already received a lot of interest in a range of image processing fields, including retinal image quality evaluation and automated retinopathy screening introduced a new retinal image quality classification (IQC) technique that merged unsupervised saliency mapping characteristics with supervised CNN features and fed them into a support vector machine: good and reduced quality retinal fundus images were automatically identified presented a residual learning framework to ease the training of a much deeper network than had previously been employed. The distributed learning architecture proposed by Chen et al. included six training layers: two fully interconnected layers and four convolutional layers that can be used to further improve the accuracy of glaucoma diagnostics. proposed a new classification-based eye disease detection method that uses deep CNN generated from a large-scale common dataset for visual representation and combines global and local features to reduce the effects of discrepancies. The road sign identification algorithm was proposed for occasions where signs were occluded by low lighting, obfuscation, loss and distortion and hence the improved VGG model (IVGG) based on the VGG model was proposed. Proposed an ALOHA-based tag identification protocol which enhanced the reading efficiency of an EPC C1 Gen2 air interface system and postulated an incremental algorithm; a rapidly evolving network policy; which overcomes the constraints on existing and previous algorithms of network software-defined architecture. Deep CNN was proposed by (add et al) to discover and subsequently identify deficiencies such as retinal exudates. The tensor structure might be utilized to improve the local pattern of edge elements and to modify the intensity of the object's edge. For the active contour approximation of picture segmentation, the gradient descent approach was employed. Introduced an interestingly new or unorthodox deep learning architecture for a set of a number of clinically diverse 3D optical coherent CT images, and (add et al part) developed a CNN based on automated deep photography images and effective computer approaches. (add the author bit) differentiated natural pictures from computerized images and put forward an efficient CNN-based technique. They proposed an approach that can adapt to varied sizes of the picture patches while retaining the depth of the fixed network, and the network was integrated into the ostensible local to the global strategy.

---

[1] * denotes equal contribution

As a result, the classification of high-frequency picture characteristics should be the primary objective of eye illness detection and classification, which may be accomplished using created forensic images, domain knowledge of natural images, and generated image forensics. The combination of the retinal fundus, retina recognition and eye disease assessment could be utilized to better correctly identify early eye illness from retinal pictures. This can be made possible when dense link blocks in DenseNet are leveraged and implemented to improve the VGG network training model to experiment on an eye disease dataset which can help significantly when trying to precisely pinpoint early eye illness from retinal pictures.

## 3. Methods

We used transfer learning as a lead role. VGG - 19 was the pre-trained model that we used to transfer its learned data to the new baseline model with a new data set. Imagenet was the data set that was used to source the data. In transfer learning, we freeze the starting layers and use the last layers as prediction layers. We added an extra sequential layer with combinations of blocks. (Explaining why we need transfer learning, talking about the parameters, pre-trained weights in terms of images, and no connection of it being with the retinal stuff. Alongside, the lack of data. Also, mention experimentation (how we tried a 3-block, 5-block and 7-block approach )

The architecture implementation that ClaRet followed was a stacking of blocks of layers. Each block represents a single set of full-fledged output, with a combination of convolutional layer, dense layer and drop-out layer. At each stage, these blocks try to gradually increase the precision rate and reduce the number of classification arguments until it reaches its final classes which in this case is 4. By stacking these multiple convolution layers we were still able to use drop-out layers to reside unnecessary neurons without losing information. These stacks of layers extract more information than just a single pair of convolutional layers. Alongside we set our Dropout layer argument to 0.3, with all experimentation this gave the best results. Since we used these layers in a block format, it becomes easier to edit the architecture according to need and just pose better coordination between layers and consequent filters. Since we gradually reduced the number of dense layers, instead of going with an ascending order of filter numbers, we made it in sync with dense layers that are descending in nature. Each set of dense layers that we added, worked as an individual output classification layer. We reduced the layers in number consequently, because then we were able to increase the level of classification in terms of class arguments at each stage of the process.

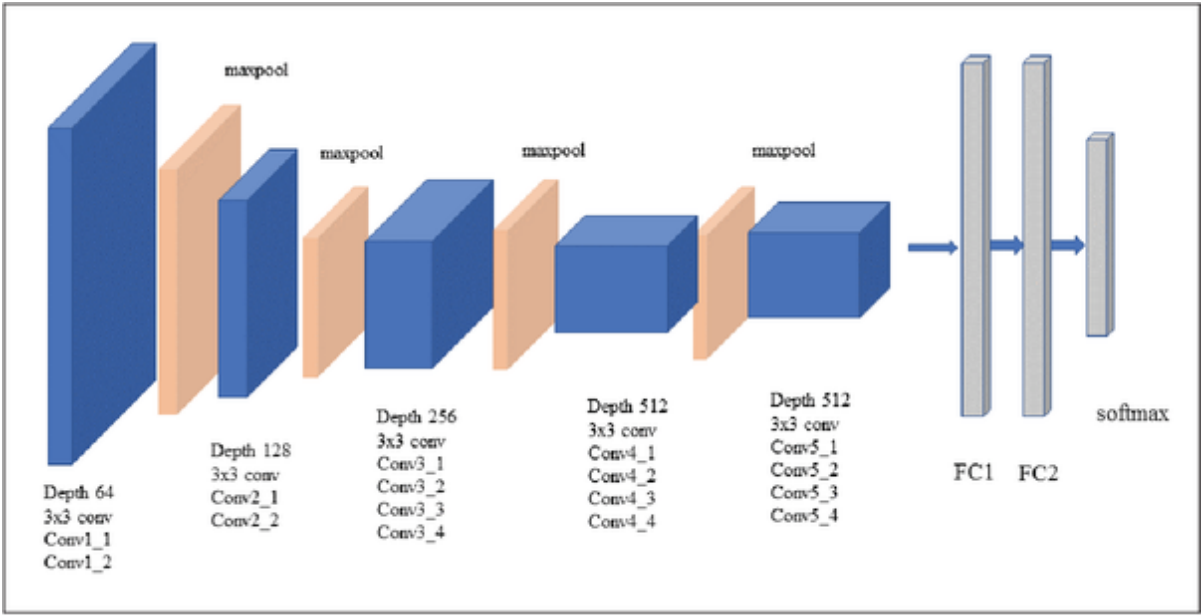

Fig. 3. VGG-19 network architecture

## 4. Architecture

This phase required a lot of experimentation, testing and brainstorming since this was the part that defined how the model interpreted its weights and data. As mentioned earlier, we had a block structure consisting of a combination of a convolutional layer, a dense layer and a dropout layer. We added our architecture parameters in the following ways for each type of layer:

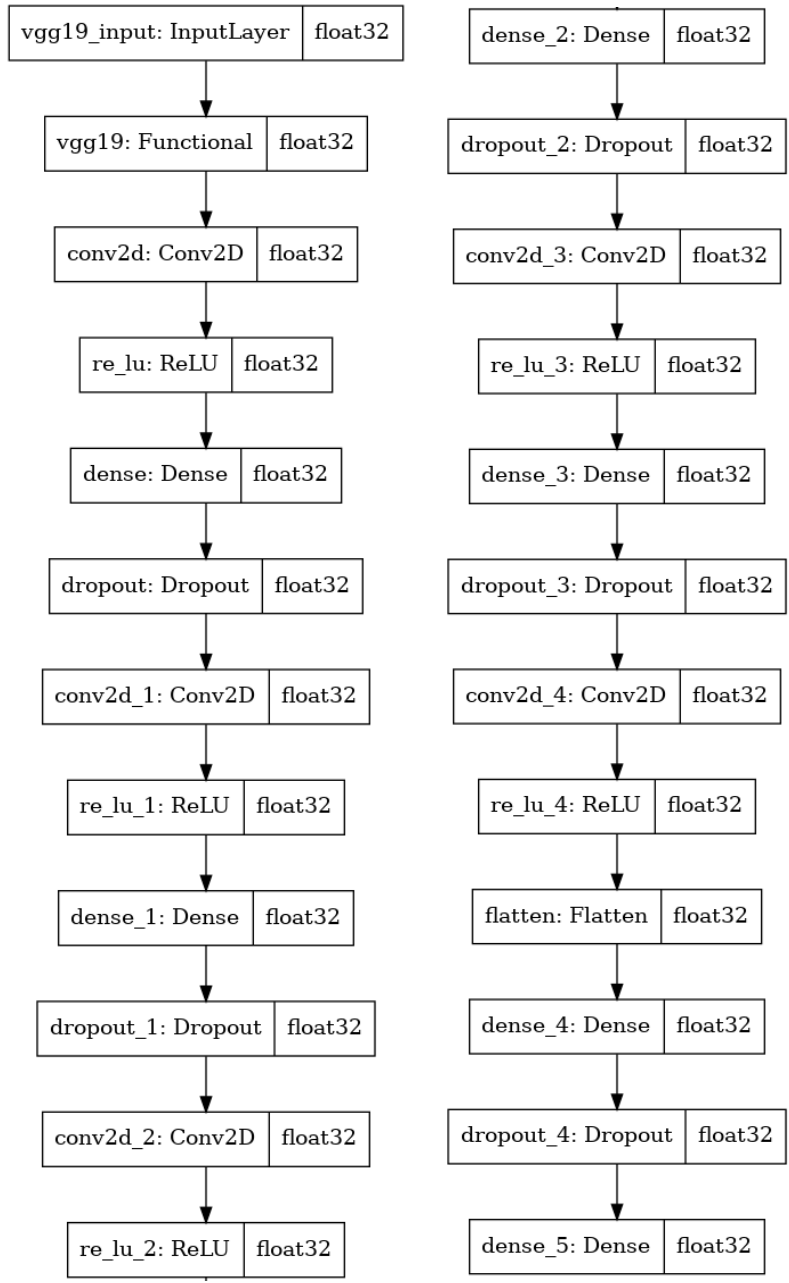

Fig X: Convolutional Neural Network Architecture

**4.1 Conv2D**

- The filters for these layers were tuned in an exponential format where the magnitude started from 16 ($2^4$) and exponentially grew to 256 ($2^8$) over a span of 5 Convolution layers.

- This was done so that the CNN could distinguish more shapes and textures within the scans to not miss out on important features

- Exponents of 2 were because the filters were directly related to the feature maps that had to be flattened at the end. This allowed the RAM of the hosting machine to organize the weights better and help in the faster fitting of the model.

- The 'SAME' parameter was given for the padding so that it was done evenly for each zero to the left/right or up/down of the input such that the output had the same height/width dimension as the input. This made it easier to fit the model without manual rescaling.

**4.2 Dense layers**

- Broadly, there were two types of Dense layers configured for the Network :
    - Dense layers with ReLU activation ()
    - Dense layers with softmax activation

- The arrangement of the values of the units for each ReLU-activated dense layer was predominantly the opposite of what we did for the Convolution layers.

- The ReLU-activated dense layer units were tuned in a logarithmic format where the magnitude started from 1024 ($2^{10}$) and logarithmically decreased to 32 ($2^5$) over a span of 5 layers.

- The units for the softmax-activated dense layer were scaled down to 4 instead of the usual 16 units as softmax is used for multivariate classification predictions and there were 4 classes present in our dataset.

**4.3 Dropout layers**

After a lot of experimentation, we decided to keep the rate for the Dropout layer equal to 0.2 as removing 20% of input units performed very well at preventing overfitting

# 5. Results

Our model achieved the following results:

| Training Accuracy | 95.73% |
| --- | --- |
| Testing Accuracy | 96.28% |
| Validation Accuracy | 96.88% |
| F1 Score | 0.9628 |

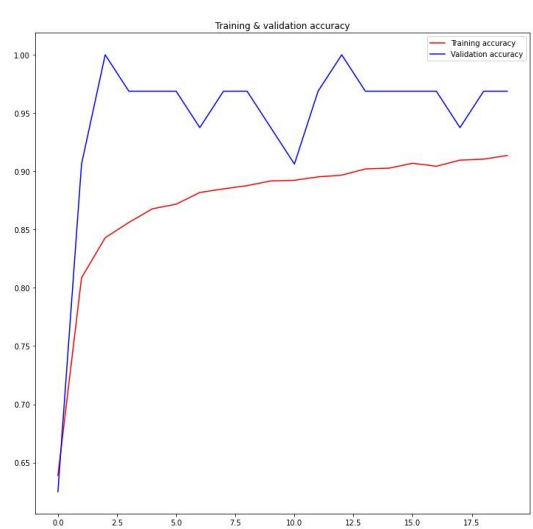 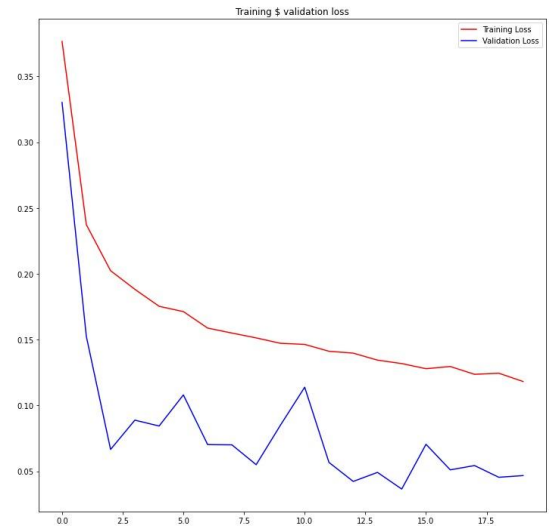

Training Accuracy   Training Loss

## 6. Conclusions

**How our general model can be applied to and help improve the performance of other tasks**

The architecture and structure of our model are very versatile and can easily be applied to other customized models by changing the Dense layer units and the Conv2D layer filters according to the weights and intensity of the model. The block structure provides an interface where the model can perform well on most datasets/ Transfer Learning models and the dropout layers to prevent overfitting. One thing to make sure of when fitting a custom model through this architecture is the change of the '4' in the fully connected layer to the number of classes in your dataset.

**Deployment of a prototype of the model on the web**
Our next step would be to try to deploy the model on the web so that a prototype can be used in real-time. We aim to do this by getting the model prediction and matching it to the class name so that it can be detected by a person and understand the actual classified prediction. Then we would try to download the model through pickle and deploy it on a prototype web interface using a framework like Streamlit / Flask / Django etc.

## References

[1]: G. Varun, P. Lily, C. Marc, M. C. Stumpe, W. Derek et al., "Development and validation of a deep learning algorithm for detection of diabetic retinopathy in retinal fundus photographs," JAMA, vol. 316, no. 22, pp.2402–2410, 2016

[2] Shah, T. (2020, July 10). About Train, Validation and Test Sets in Machine Learning - Medium. https://towardsdatascience.com/train-validation-and-test-sets-72cb40cba9e7

[3] Tsuji, T., Hirose, Y., Fujimori, K. et al. Classification of optical coherence tomography images using a capsule network. BMC Ophthalmol 20, 114 (2020). https://doi.org/10.1186/s12886-020-01382-4

[4] Wang, Depeng & Wang, Liejun. (2019). On OCT Image Classification via Deep Learning. IEEE Photonics Journal. PP. 1-1. 10.1109/JPHOT.2019.2934484.

[5] Hui, Vivian W.K. MBChB*,†; Szeto, Simon K.H. FCOph(HK), FHKAM (Oph)*,†; Tang, Fangyao PhD*; Yang, Dawei MBBS, MMed*; Chen, Haoyu MD‡; Lai, Timothy Y.Y. MD, FRCOphth*,§; Rong, Ao MD¶,‖; Zhang, Shaochong MD, PhD**; Zhao, Peiquan MD, PhD††; Ruamviboonsuk, Paisan MD‡‡; Lai, Chi-Chun MD§§; Chang, Andrew PhD, FRANZCO¶¶; Das, Taraprasad MD‖‖; Ohji, Masahito MD, PhD***; Huang, Suber S. MD, MBA†††,‡‡‡; Sivaprasad, Sobha DM, FRCOphth§§§; Wong, Tien Yin MD, PhD¶¶¶,‖‖‖; Lam, Dennis S.C. MD****,††††; Cheung, Carol Y. PhD*. Optical Coherence Tomography Classification Systems for Diabetic Macular Edema and Their Associations With Visual Outcome and Treatment Responses – An Updated Review. Asia-Pacific Journal of Ophthalmology: June 2022 - Volume 11 - Issue 3 - p 247-257
doi: 10.1097/APO.0000000000000468